\documentclass{mva_style}
\usepackage{amsmath}
\usepackage{graphicx}
\usepackage{here}
\usepackage{booktabs}
\usepackage{colortbl}
\usepackage{xcolor}
\usepackage{tabularx}
\usepackage{afterpage}
\finalcopy 

\newcommand{\argmax}{\mathop{\rm arg~max}\limits}
\newcommand{\argmin}{\mathop{\rm arg~min}\limits}

\usepackage{titlesec}
\titlespacing*{\chapter}{0pt}{50pt}{40pt}

\begin{document}
\title{Leveraging 2D-VLM for Label-Free 3D Segmentation \\
in Large-Scale Outdoor Scene Understanding}

\author{
  Toshihiko Nishimura, Hirofumi Abe, Kazuhiko Murasaki, Taiga Yoshida, Ryuichi Tanida\\
  NTT Corporation
}

\maketitle

\section*{\centering Abstract}
\textit{
This paper presents a novel 3D semantic segmentation method for large-scale point cloud data that does not require annotated 3D training data or paired RGB images. The proposed approach projects 3D point clouds onto 2D images using virtual cameras and performs semantic segmentation via a foundation 2D model guided by natural language prompts. 3D segmentation is achieved by aggregating predictions from multiple viewpoints through weighted voting. Our method outperforms existing training-free approaches and achieves segmentation accuracy comparable to supervised methods. Moreover, it supports open-vocabulary recognition, enabling users to detect objects using arbitrary text queries—thus overcoming the limitations of traditional supervised approaches.
}

\afterpage{
\begin{figure*}[t]
    \centering
    \includegraphics[width=0.90\linewidth]{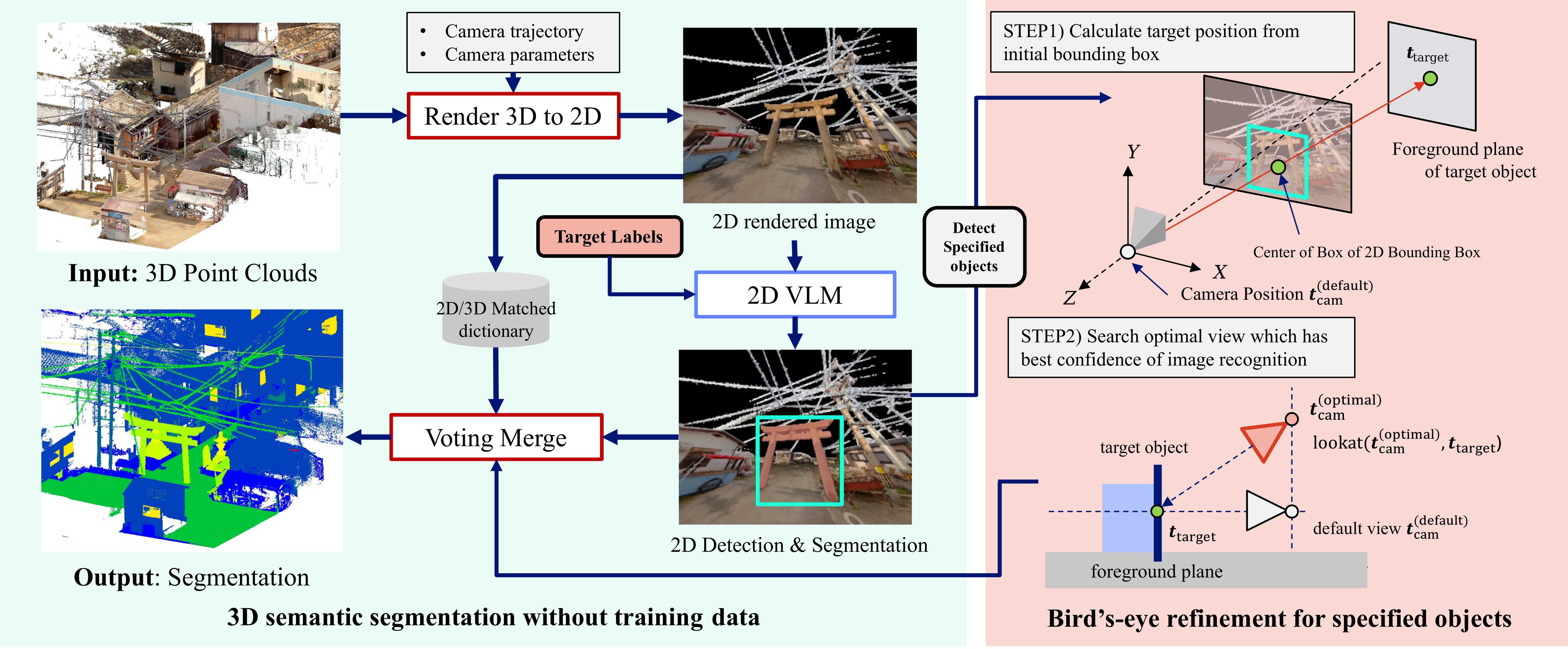}
    \caption{\textbf{An overview of the proposed method.} 3D points are projected into 2D views using virtual cameras. A 2D-VLM segment targets and labels are fused back into 3D by voting, with optional bird's-eye refinement module.}
    \label{fig:overview}
\end{figure*}
}

\section{Introduction}
3D scene understanding has become increasingly important with the growing availability of sensors such as depth cameras and LiDAR devices, which enable the acquisition of rich 3D visual information. 3D information plays a significant role in various applications, including autonomous driving, extended reality and construction. Calibrated cameras often provide RGB color aligned with 3D point clouds, while metadata such as semantic segmentation labels is required for downstream tasks like simulation and digital twin generation. As a result, 3D scene understanding has been studied to support the development of intelligent systems across various domains.

Deep neural architectures for processing point clouds have been actively studied for many years \cite{PointNet, PointNet++, DGCNN, kpconv, RandLANet, PTv1, PTv2, PTv3}. Most of these methods address closed-set segmentation tasks, which aim to recognize a predefined set of semantic labels. When sufficient annotated data and computational resources are available, they achieve strong performance on various point cloud benchmarks. However, in practical scenarios, the cost of collecting and annotating large-scale 3D point cloud datasets and training deep models is prohibitively expensive. Additionally, the characteristics of point cloud data—such as scan density, calibration accuracy, and scanning range—can vary significantly depending on the type of measurement device and reconstruction algorithm used. These factors make supervised learning for point clouds significantly more expensive and labor-intensive compared to the image and language domains, where large-scale annotated datasets are more readily available.
In contrast, the image and language domains have greatly benefited from internet-scale datasets that have enabled the development of general-purpose models and open-vocabulary recognition\cite{CLIP, OpenSeg, Lseg, chen2023open}. Achieving similar capabilities for 3D point clouds remains challenging due to the difficulty of collecting such massive 3D data. There are various methods that leverage intermediate 2D images to link 3D data with language\cite{CLIP2Scene, OpenMask3D, PLA, RegionPLC, OpenIns3D}; they often rely on numerous RGB images that are often discarded to save storage in practice. Moreover, these approaches have been demonstrated only on indoor scenes or within limited outdoor areas.

In this paper, we propose a method for semantic segmentation of wide-area LiDAR point clouds by leveraging a 2D vision model. Without requiring any annotations or training, images are rendered along the LiDAR trajectory and segmented in 2D. The results from multiple virtual views are then projected back and fused into 3D space via a voting scheme. This approach enables open-vocabulary segmentation in large-scale outdoor environments. Experimental results demonstrate that our method outperforms existing training-free approaches and achieves segmentation performance approaching that of fully supervised methods.

\section{Related Work}

\textbf{Supervised Approach.} 
The segmentation of 3D point clouds has been studied for a long time in the fields of computer vision and robotics, even prior to the advent of deep learning architectures for point clouds\cite{MVCNN, Shapenet}. 
Since the introduction of PointNet, numerous architectures have been developed to learn from point cloud data and 3D ground-truth labels, including point cloud convolution methods\cite{DGCNN, kpconv}, efficient handling of wide-area point clouds\cite{RandLANet}, and Transformer-based models that improves accuracy\cite{PTv1, PTv2, PTv3}. 
Point cloud data varies in scale and context, ranging from individual object scans to 2.5D data captured by autonomous vehicles and robots, as well as large-scale scenes aggregated from multiple scans. Corresponding benchmark datasets have been proposed for each of these scenarios \cite{SensatUrban, Semantic3D, KITTI, nuScenes}.
However, supervised learning is often prohibitively expensive in practice, due to the high cost of collecting and annotating ground-truth data. 
Furthermore, supervised models often struggle to accurately predict infrequent classes or fine-grained objects due to the inherent class imbalance in point cloud datasets. 
The approach proposed in this paper addresses these challenges by leveraging an image-based model.

\textbf{Leveraging 2D vision models.}

To mitigate the challenges associated with supervised learning for 3D point clouds, recent studies have increasingly explored the use of foundation models originally developed for image understanding.
In particular, image-language models such as CLIP have attracted considerable attention, as they enable generic tasks like classification\cite{PointCLIP, PointCLIPv2, CLIP2Point}, object detection \cite{OV3DET, OV3DETR}, and segmentation\cite{PLA, RegionPLC, OpenMask3D, Open3DIS, OpenIns3D} based on arbitrary language queries.
However, most of these studies have focused on limited scenarios, such as indoor environments or temporally varying LiDAR scans, and their effectiveness in large-scale outdoor scenes remains largely unverified.
In addition, many approaches assume the availability of RGB images paired with point clouds. 
In practice, however, such images are often discarded to save storage, with only colorized point clouds being retained, making these methods difficult to apply.
In this work, we propose a method that enables the segmentation of wide-area outdoor point clouds by utilizing rendered images.

\section{Methods}

The pipeline of the proposed method is illustrated in Figure 1. 
The green hatched block represents the segmentation process, which operates without training. 
Given an input point cloud, a camera trajectory, and rendering parameters, projected images are generated. A 2D vision-language model (VLM) is then applied to perform recognition on these images.
The segmentation results from multiple views are aggregated onto the corresponding 3D points using a voting-based approach, producing the final 3D segmentation output.
The red hatched block indicates an optional module that performs vertical camera shifting and refinement when a specific object is detected.
This module is designed to enhance recognition accuracy from a bird’s-eye perspective, particularly for objects that are difficult to recognize reliably from the default viewpoint.
Each component is described in detail below.

\textbf{2D Projection and Segmentation.}  A virtual camera with orientation $R$ is placed at position $\boldsymbol{t}$ in the 3D point cloud space. 
Given the intrinsic parameter matrix $K$ of the virtual camera, the 3D points are projected onto the image plane using the camera coordinate system $\boldsymbol{u} = [u,v]^\top$.
The world coordinates $\boldsymbol{x}_{\mathrm{world}}$ are transformed into image coordinates via the equation $\boldsymbol{u} = K [R | \boldsymbol{t}] \cdot \boldsymbol{x}_{\mathrm{world}}$. A mapping between 2D pixels and 3D points is maintained as a dictionary for later use.
2D semantic segmentation is performed on the rendered images using a 2D vision-language model (2D-VLM).
In our approach, Grounded SAM\cite{GroundedSAM}, which combines GroundingDINO\cite{GroundingDINO} and Segment Anything Model (SAM)\cite{SAM}, is applied here. First, a list of object class names is provided as input queries of GroundingDINO, and the detection rectangle containing the object class names is output. Next, the resulting semantic labeled rectangle is input to SAM to obtain a segmentation mask. The semantic label associated with each rectangle is assigned to the corresponding segmentation mask extracted by SAM, and the result is output as a semantic segmentation result.

\textbf{Weighted Voting.} The method for integrating point clouds labeled by virtual cameras into the original large-scale point cloud is illustrated in Figure~\ref{fig:voting-merge}. Let $P = \{p_n\}_{n=1}^{N}$ denote the $N$ original point cloud, and let $Q^{(c)} = \{(q_m^{(c)}, l_m^{(c)})\}_{m=1}^{M^{(c)}}$ represent a partial $M^{(c)}$ point cloud captured by the $c$-th camera, where each point $q_m^{(c)}$ is associated with a label $l_m^{(c)}$. For each point $p_n \in P$, we search for its neighboring labeled points within a distance threshold $\varepsilon$ from all partial point clouds $Q^{(c)}$, and collect them into a set $\mathcal{N}(p_n)$:

{
\setlength{\abovedisplayskip}{0pt}
\begin{multline}
    \mathcal{N}(p_n)=\left\{ \argmin_{q_m^{(c)} \in Q^{(c)}} \| p_n - q_m^{(c)} \| \,\middle|\, \right. \\
     \left. \| p_n - q_m^{(c)} \| < \varepsilon,\; c=1,2,\dots,N_{\mathrm{cam}} \right\}
\end{multline}
}

For each point in $P$, a label vote is cast using neighboring labels weighted by both the recognition confidence $w_n^{(c)}$ and the inverse of the distance to the camera $1/d_n^{(c)}$, as defined in Eq.~(\ref{eq:voting}):

\begin{equation}
    V_l(p_n) = \sum_{\mathcal{N}(p_n)}   \dfrac{w^{(c)}_n}{d_n^{(c)}} \delta ( l, l^{(c)}_n )     \label{eq:voting}
\end{equation}

Finally, the label with the highest accumulated weight is assigned to the point:

\begin{equation}
    l_n = \argmax_l V_l(p_n)
\end{equation}

\begin{figure}[b]
    \centering
    \includegraphics[width=1.0\linewidth]{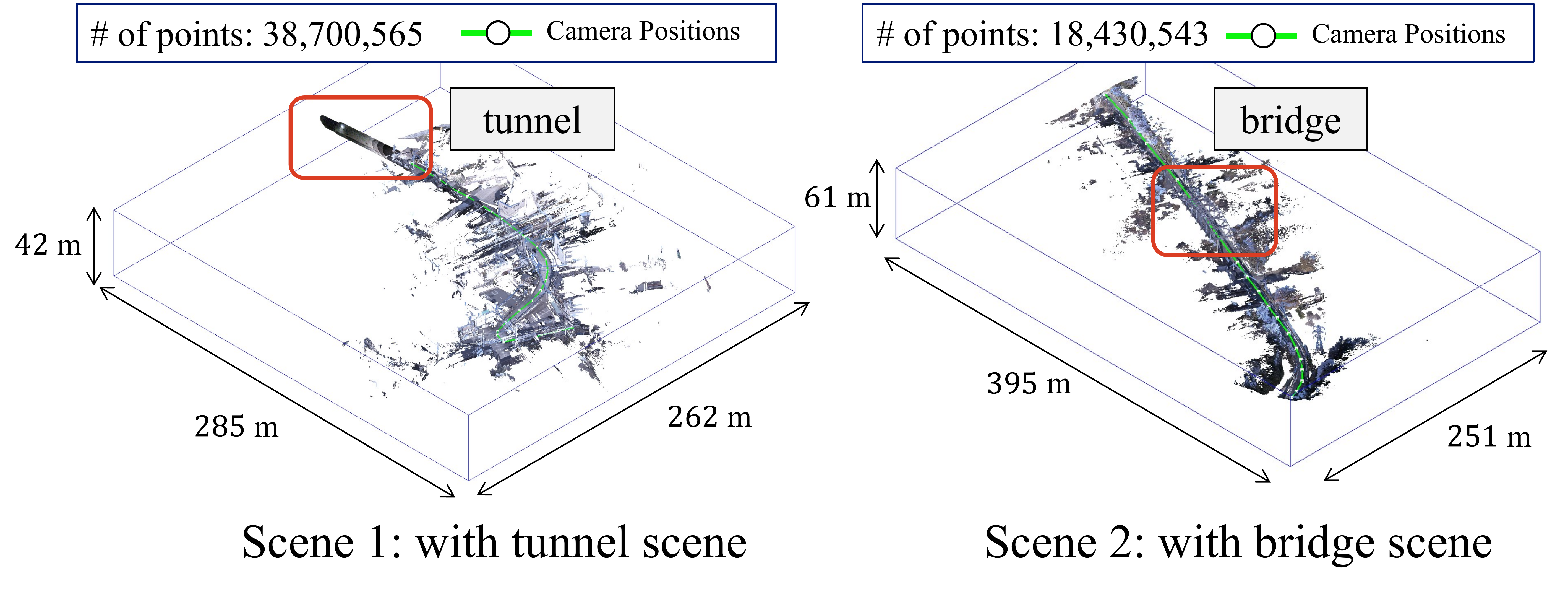}
    \caption{\textbf{Test scenes used in our experiments.} Scene 1 contains a tunnel, and Scene 2 contains a bridge.}
    \label{fig:enter-label}
\end{figure}

\afterpage{
\begin{figure*}[h]
    \centering
    \includegraphics[width=0.85\linewidth]{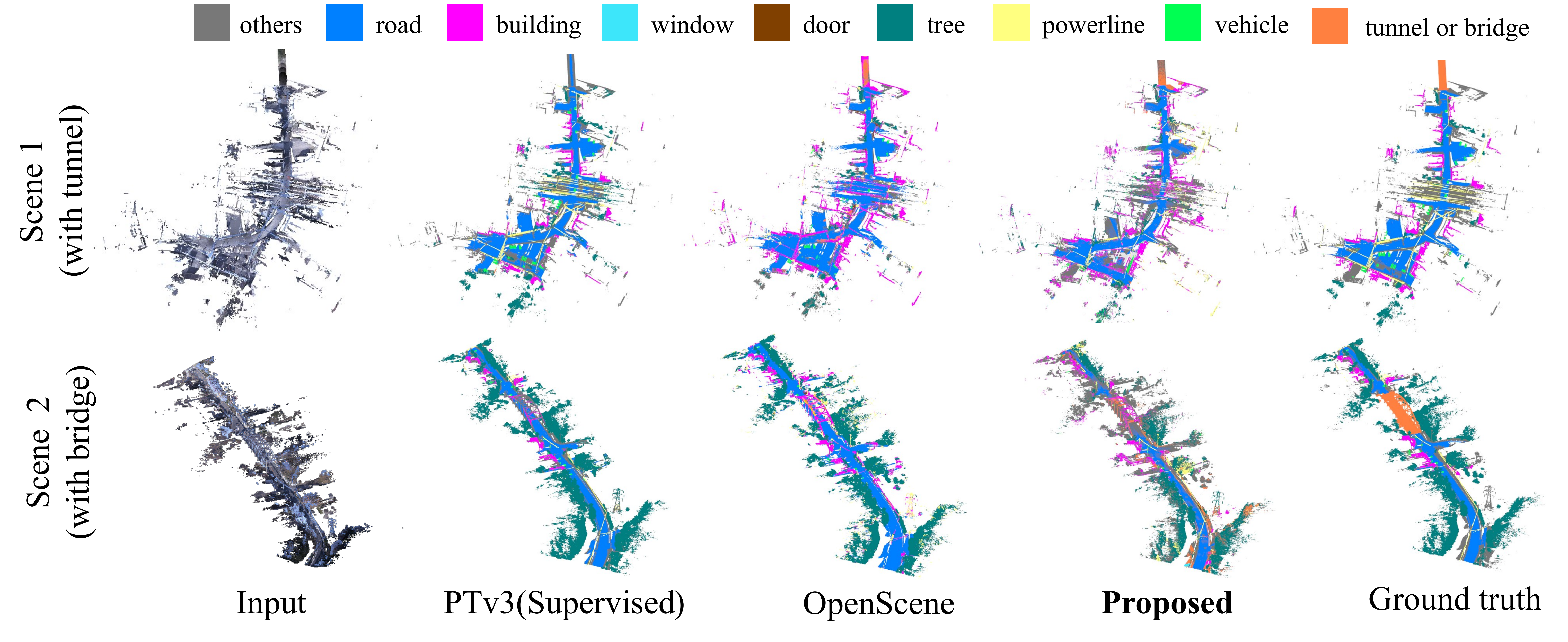}
    \caption{\textbf{Qualitative comparisons.} Images of 3D segmentation results on our test dataset.}
    \label{fig:segmentation_result}
\end{figure*}
}
\begin{figure}
    \centering
    \includegraphics[width=0.8\linewidth]{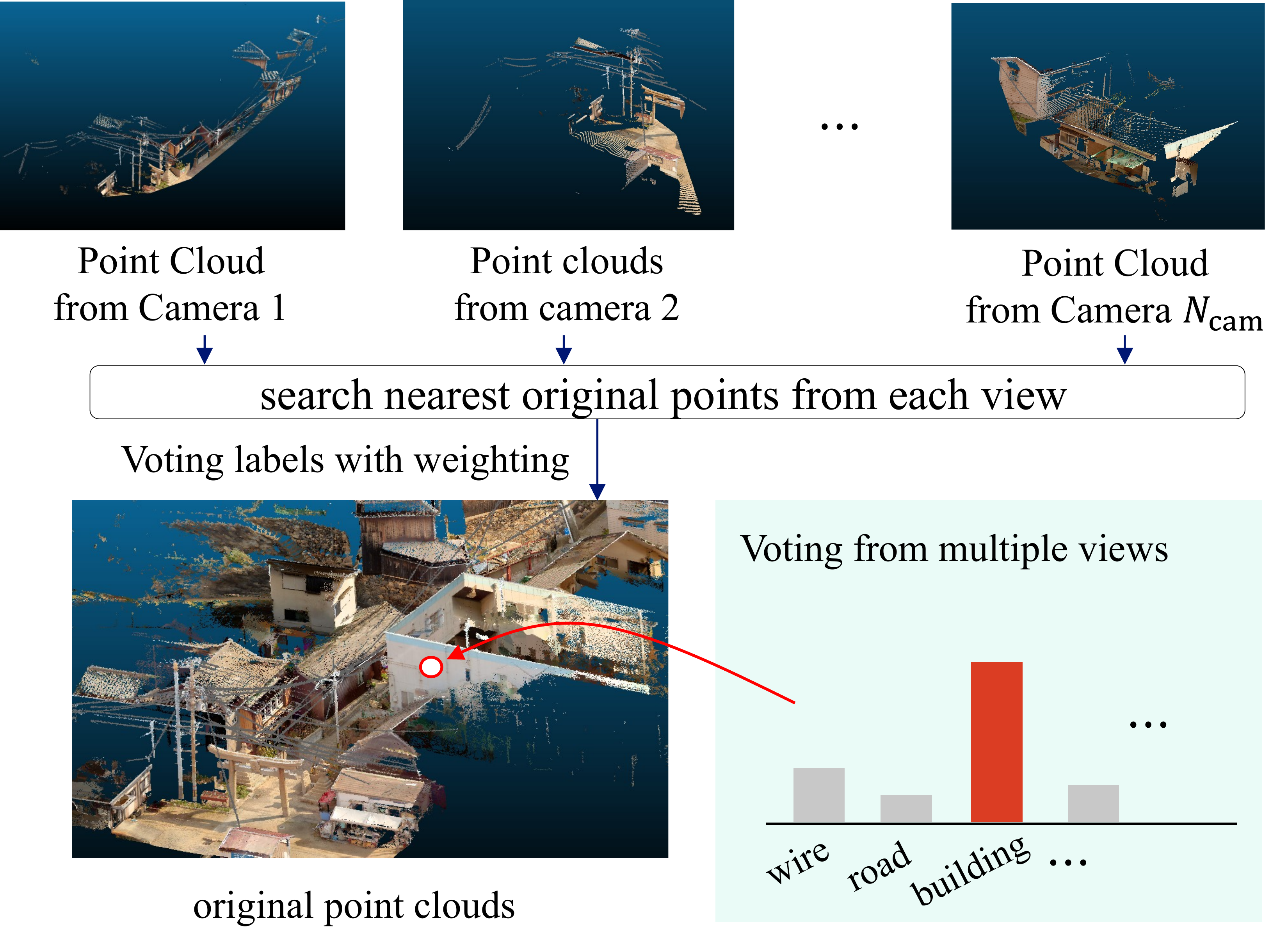}
    \caption{\textbf{Voting Merge.} Point clouds from virtual cameras are weighted by confidence score and distance from the camera, then voted on. The highest-scoring label becomes the final result.}
    \label{fig:voting-merge}
\end{figure}

\textbf{Bird's-eye Refinement.}
The camera trajectory follows the driving path during data acquisition. While many objects can be recognized from a ground-level view, large or elongated structures are often more reliably identified from a bird’s-eye perspective. Therefore, when a user-specified object is detected along the trajectory, the camera switches to a top-down viewpoint to improve recognition accuracy. The object is then re-rendered and re-recognized from this new perspective, and the 2D segmentation result with the highest confidence is projected onto the 3D point cloud. The camera pose is computed using a standard look-at transformation, which targets the point where the center of the bounding box in the initial view intersects with the front surface of the object.

\begin{table}[t]
\caption{Comparison Performance}
\label{tab:zeroseg_performance}
\centering
\renewcommand{\arraystretch}{1.3}
\scalebox{0.8}{
\begin{tabularx}{\linewidth}{c *{2}{>{\centering\arraybackslash}X}}
\hline
 & Scene1(mIoU) & Scene2(mIoU) \\
\rowcolor[gray]{0.9}
PTv3(supervised)\cite{PTv3} & 0.436 & 0.441 \\
OpenScene\cite{OpenScene} & 0.329 & 0.335 \\
\textbf{Proposed} & \textbf{0.397} & \textbf{0.375} \\
\hline
\end{tabularx}
}

\caption{Bird’s-eye Refinement Performance}
\label{tab:refined_performance}
\centering
\renewcommand{\arraystretch}{1.3}
\scalebox{0.8}{
\begin{tabularx}{\linewidth}{c *{2}{>{\centering\arraybackslash}X}}
\hline
 & Tunnel(IoU) & Bridge(IoU) \\
\rowcolor[gray]{0.9}
PTv3(supervised)\cite{PTv3} & 0.000 & 0.000 \\
OpenScene\cite{OpenScene} & 0.134 & 0.031 \\
Proposed & \underline{0.675} & \underline{0.130} \\
\textbf{Proposed(refined)} & \textbf{0.695} & \textbf{0.397} \\
\hline
\end{tabularx}
}

\end{table}

\section{Experimental Results}

\textbf{Datasets.}
We evaluated the proposed method using a 3D point cloud dataset acquired by a camera and LiDAR mounted on a Mobile Mapping System (MMS).
To compute recognition accuracy, we manually annotated seven object categories: road, building, window, door, powerline, vehicle, and tree, along with two rare instance-level objects: tunnel and bridge.
The proposed method was tested on two scenes: Scene 1, which includes a tunnel, and Scene 2, which includes a bridge.

\textbf{Setup.}
The travel paths recorded during data acquisition were used as trajectories for placing virtual cameras in each dataset.
Each virtual camera was configured with a 90-degree field of view, an image resolution of 640 pixels in height and 480 pixels in width.
Rendered point clouds were visualized by displaying each point as a sphere with a radius of 0.01 meters.
Bird’s-eye refinement was applied to Scene 1 (tunnel) and Scene 2 (bridge) as these objects were designated for enhanced recognition.
When any of the target objects were detected, the camera was vertically shifted by 5 m, 10 m, and 15 m from the initial position, and the 2D segmentation result with the highest recognition confidence was selected.

\textbf{Comparison Performance.} In recent years, many Zero-Shot segmentation methods have been proposed. OpenMask3D\cite{OpenMask3D} and CLIP2Scene\cite{CLIP2Scene} are for range-limited areas and require pre-training to adapt our dataset. Hence, we compare OpenScene\cite{OpenScene}, which can infer a wide range of point clouds from pre-trained models. We used a pretrained model that distilled 2D features from the nuScenes dataset without requiring additional training data.
Table~\ref{tab:zeroseg_performance} reports mIoU results. In both scenes, the proposed method outperforms the baseline.
To examine the performance gap with a supervised approach, we evaluated the Point Transformer V3 (PTv3) model trained on scenes other than Scene 1 and Scene 2. As expected, the supervised model achieved higher accuracy. However, our method performs comparably while requiring no annotated data, as illustrated in Fig.~\ref{fig:segmentation_result}.
Table~\ref{tab:refined_performance} shows IoU scores for rare objects—tunnel and bridge—in Scene 1 and Scene 2, respectively. The supervised model failed to detect them due to limited training samples, whereas both the baseline and our method succeeded. Bird’s-eye refinement further improved accuracy for these classes.

\begin{figure}[t]
    \centering
    \includegraphics[width=0.85\linewidth]{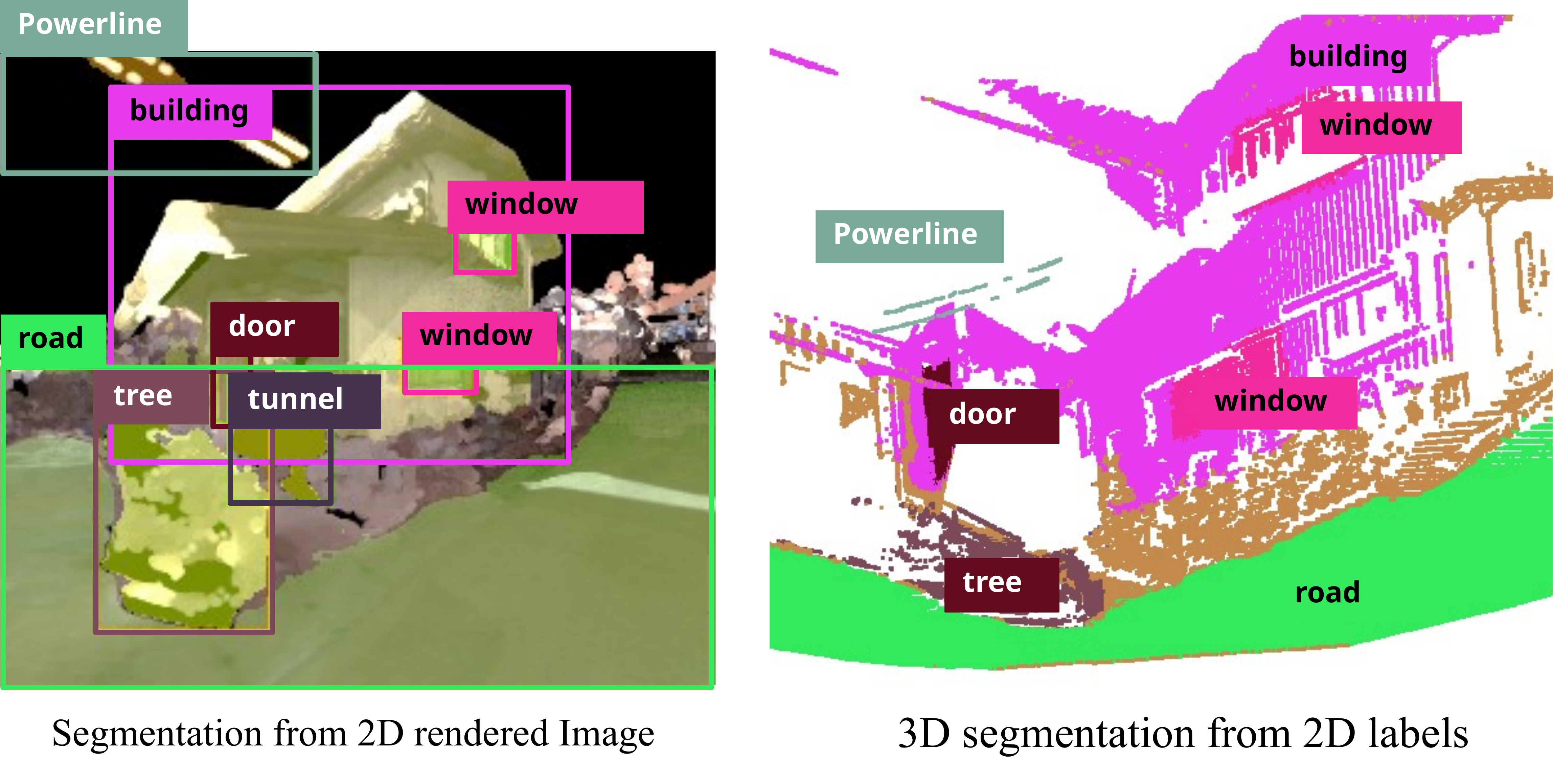}
    \caption{\textbf{Matched Results from 2D to 3D.} 2D segmentation results are transferred to 3D point clouds}
    \label{fig:Matching2Dto3D}
\end{figure}

\begin{figure}[t]
    \centering
    \includegraphics[width=0.85\linewidth]{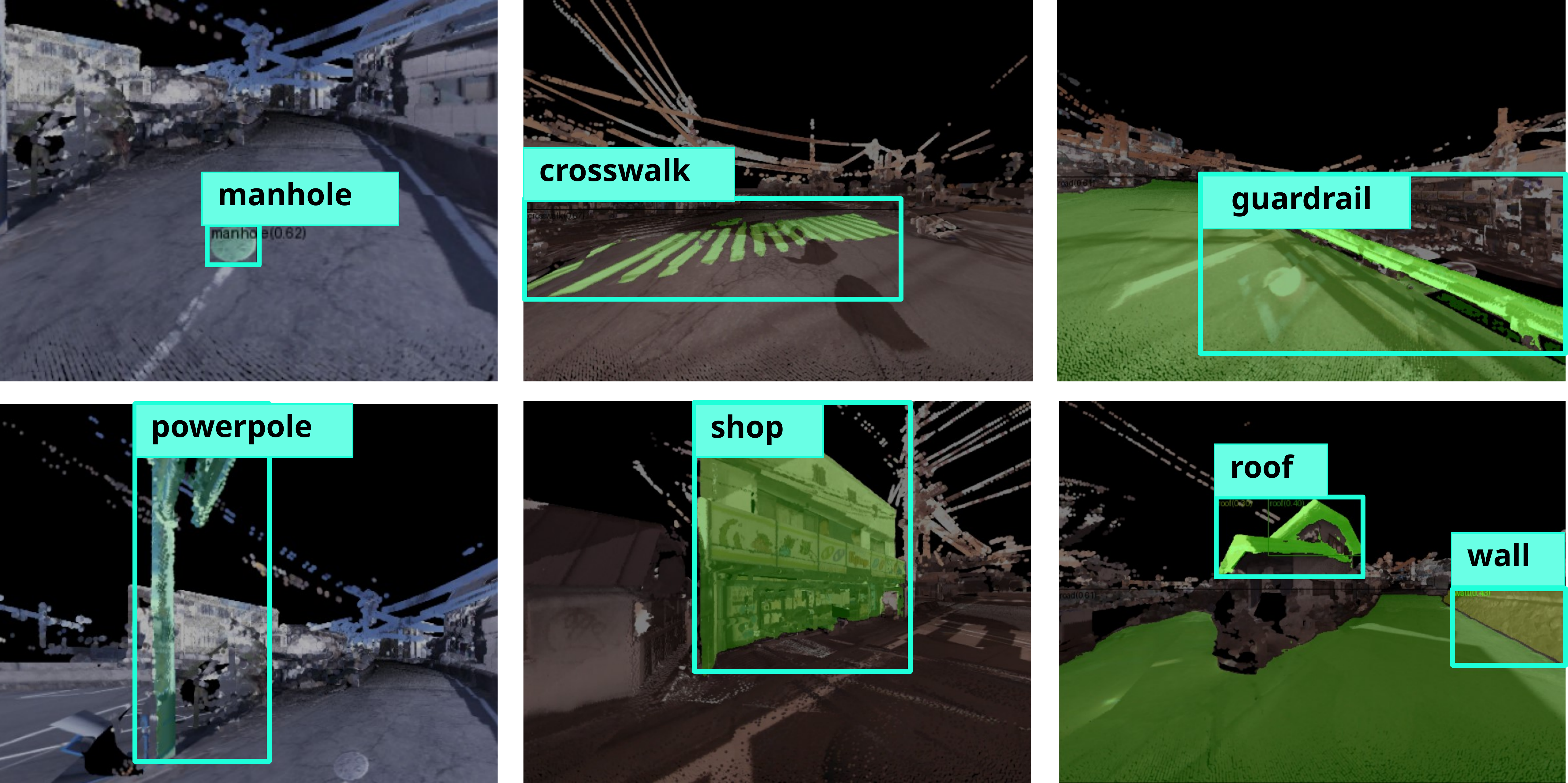}
    \caption{\textbf{Open vocabulary segmentation.} Arbitrary objects can be recognized with 2D-VLM applied to 2D rendered images.}
    \label{fig:ov_seg}
\end{figure}

\textbf{Qualitative Segmentation Performance.}
Figure~\ref{fig:Matching2Dto3D} illustrates the correspondence between the segmentation result and the 3D point cloud rendered from MMS data. Objects such as buildings, windows, and doors are correctly segmented from the projected image of the point cloud. Although some regions without point cloud data are incorrectly labeled as tunnels, these false positives do not affect the recognition of actual 3D points, as they occur in areas where no 3D data exists.
Figure~\ref{fig:ov_seg} shows examples of labels that were not included in the dataset but were still detected by the model when given as queries. Rare object classes such as manholes and pedestrian crossings are typically difficult to annotate in supervised learning due to the scarcity of training samples. However, the use of vision-language models enables recognition of such rare objects without requiring additional supervision.

\section{Conclusion}
This paper proposes a method for semantic segmentation of large-scale 3D point clouds by projecting them into 2D images and applying image-based recognition. Since 2D images are not required at inference time, the method is suitable for cases with only colorized point clouds, such as synthetic CG data. Currently, it uses predefined trajectories with limited viewpoints. Future work will explore data-driven camera placement and integration with 3D structure understanding.

\bibliographystyle{unsrt}
\bibliography{reference}

\end{document}